\documentclass{article}

\usepackage{PRIMEarxiv}

\usepackage[utf8]{inputenc} 
\usepackage[T1]{fontenc}    
\usepackage{hyperref}       
\usepackage{url}            
\usepackage{booktabs}       
\usepackage{amsfonts}       
\usepackage{nicefrac}       
\usepackage{microtype}      
\usepackage{lipsum}
\usepackage{fancyhdr}       
\usepackage{graphicx}       
\graphicspath{{media/}}     
\usepackage{multicol}
\usepackage{amssymb}
\usepackage{graphicx}%
\usepackage{multirow}%
\usepackage{amsmath,amssymb,amsfonts}%
\usepackage{amsthm}%
\usepackage{mathrsfs}%
\usepackage[title]{appendix}%
\usepackage{xcolor}%
\usepackage{textcomp}%
\usepackage{manyfoot}%
\usepackage{booktabs}%
\usepackage{algorithm}%
\usepackage{algorithmicx}%
\usepackage{algpseudocode}%
\usepackage{listings}%
\usepackage[justification=centering]{caption}%
\usepackage{makecell}
\usepackage{CJKutf8}
\title{Iterative missing value imputation based on feature importance

}

\author{
  Cong Guo, Chun Liu,Wei Yang* \\
  Henan University \\
  China \\
  Kaifeng\\
  \texttt{guocong@henu.edu.cn,liuchun@henu.edu.cn,yangwei@henu.edu.cn} \\
}

\begin{document}
\maketitle

\begin{abstract}
Many datasets suffer from missing values due to various reasons,which not only increases the processing difficulty of related tasks but also reduces the accuracy of classification. To address this problem, the mainstream approach is to use missing value imputation to complete the dataset. Existing imputation methods estimate the missing parts based on the observed values in the original feature space, and they treat all features as equally important during data completion, while in fact different features have different importance. Therefore, we have designed an imputation method that considers feature importance. This algorithm iteratively performs matrix completion and feature importance learning, and specifically, matrix completion is based on a filling loss that incorporates feature importance. Our experimental analysis involves three types of datasets: synthetic datasets with different noisy features and missing values, real-world datasets with artificially generated missing values, and real-world datasets originally containing missing values. The results on these datasets consistently show that the proposed method outperforms the existing five imputation algorithms.To the best of our knowledge, this is the first work that considers feature importance in the imputation model.
\end{abstract}


\section{Introduction}
Datasets are an essential core of classification or regression tasks\cite{hasan2021missing}. However, in reality, many real-world datasets contain a certain proportion of missing values, with 45\% of datasets in the UCI database having missing values\cite{aittokallio2010dealing}. Missing data makes many mathematical analysis methods impractical,for example,a core of many algorithms:the distances between samples will become impossible to calculate; thus, handling missing values is an important research topic \cite{pearl2013recoverability,rubin1976inference}. The difficulty of handling missing values depends on the nature of the missing mechanism\cite{pearl2013recoverability}, so it is necessary to detect the missing mechanism of the data before handling missing values.

The missing mechanism reflects the relationship between missing data and variable values in the dataset, thus explaining the reason for missing data. In 1976, Rubin et al. divided the missing mechanism into three types: Missing Completely At Random (MCAR), Missing At Random (MAR), and Missing Not At Random (MNAR) \cite{rubin1976inference}. MCAR assumes that missing data is independent of observed values.In a certain sense, the observed data of a dataset under MCAR can be regarded as a purely random sample of the complete data, whose mean, variance and overall distribution are similar to the complete observed data; therefore, datasets under MCAR has better classification accuracy than those under MNAR and MAR\cite{seijo2019biases}. MAR is more common than MCAR and indicates that missing data does not depend on the missing values themselves, such as in a social survey where the degree of income missingness may vary with the age of the interviewee. MNAR indicates that missing data depends on unobserved data, such as when people with high or low income refuse to answer questions, resulting in MNAR.

A large body of research has been devoted to strategies for handling missing values, among which imputation is an effective method that uses observed values to estimate the missing values. Therefore, the quality of imputed values is crucial for training predictive models and achieving accurate results \cite{samad2022missing}. Literature \cite{farhangfar2008impact} shows that common imputation methods can improve classification accuracy when the missing rate exceeds 10\%. However, most imputation methods treat each feature in the dataset as equally important, whereas in reality, there are often many redundant features in the dataset, and only a small subset of features are relevant\cite{urbanowicz2018relief}. These methods ignore the importance of features and their impact on downstream tasks, making it difficult to guide the selection of feature subsets and the construction of classifiers.

To address this issue, this paper proposes an iterative imputation method that can use the feature weights to complete the dataset with missing values, which is divided into a matrix completion stage (M-stage) and a feature weight learning stage (W-stage), to improve the imputation quality of important feature items. Experimental results show that the proposed method produces imputed datasets that are more conducive to learning important features with feature selection algorithms and achieve better classification performance.

The remainder of this paper is organized as follows: Section 2 reviews related work, Section 3 provides a detailed description of the proposed algorithm, Section 4 presents the experimental results on both real-world and synthetic datasets, and finally, the conclusion of this paper is given in Section 5.

\section{Raleted work}
\subsection{Handling of missing value}
When there are only a few missing values in a dataset, samples containing missing values can be directly deleted without significantly affecting the final analysis results \cite{raymond1987comparison,strike2001software}. However, directly deleting samples when the number of missing values exceeds a certain threshold can lead to a significant deterioration in classification performance \cite{quinlan1989unknown}. To handle this situation, some studies attempt to model the data without imputation. For example, paper \cite{tran2016improving} proposes a wrappered feature selection method that uses particle swarm optimization (PSO) combined with C4.5 to search for feature subsets. C4.5 can build decision trees under the influence of missing values because it introduces the missing situation into the calculation of information gain ratio, which can calculate the optimal subset partition of each attribute. In addition, \cite{doquire2012feature} proposes a method that uses partial distance strategy (PDS) to find the nearest neighbors for each sample to estimate the mutual information value of the datasets with missing values. The partial distance strategy is a similarity measure that only calculates the distance between known items of samples. Although the methods of ignoring missing values have been successfully applied to various problems, they do not take into account the rich information between missing values and observed values, especially when there are a large number of missing values in the dataset, the accuracy of the methods of ignoring missing values will be greatly affected in such cases.

On the other hand, imputation methods are more commonly used to deal with missing values \cite{lin2022deep,rubin2019statistical}. Among statistical-based missing value imputation methods, the most commonly used imputation technique is Expectation-Maximization (EM) imputation, which is a powerful maximum likelihood estimation method that consists of two steps, E-step and M-step, in each iteration until convergence. EM imputation has been successfully applied to various datasets with missing values in literature \cite{polikar2010learn++,zhu2012robust}. Among machine learning based imputation techniques, K-nearest neighbor (KNN) imputation is the most popular \cite{lin2020missing}. It first finds the K nearest neighbors of the sample with missing value and fills the missing part with the average value on that feature. The computation of KNN imputation is usually expensive \cite{garcia2010pattern} because it needs to calculate neighbors for each sample. In addition, deep learning models have also extended missing value imputation models, and imputation methods based on deep generative models \cite{camino2019improving}, convolutional neural networks \cite{zhuang2019innovative}, and recurrent neural networks \cite{zhang2018missing} have been developed. These imputation methods have been proven effective to some extent. However, we found that most existing studies have ignored the impact of imputation methods on downstream tasks and have not paid attention to the influence of feature importance on imputation effect. Therefore, this has become the focus of this study.
\subsection{Feature selection}
Feature selection is an important preprocessing step in machine learning, which aims to select the most discriminative features from the raw data while removing irrelevant features, often leading to improved model performance and reduced complexity. Feature selection can be roughly divided into three categories: filter, wrappered, and embedded methods. Filter-based feature selection is a type of statistical testing based method, which is independent of any specific classifier and evaluates the importance of features based on their correlation with the target variable \cite{raymond1987comparison}. Common criteria for filter-based feature selection include mutual information \cite{li2017granular,zhou2022feature} and correlation coefficient \cite{liu2020daily}, which are popular in practice due to their simplicity and speed. However, the selected features may not be optimal as there is a lack of specific learning algorithms guiding the feature selection process \cite{li2017feature}.Wrappered feature selection is a method based on machine learning classifiers, which evaluates the importance of features based on the performance of the classifier. Common models for wrappered methods include particle swarm optimization \cite{xue2014particle} and ant colony optimization \cite{ghimatgar2018improved}. The drawback of wrappered methods is that they are computationally expensive, and the quality of the selected feature subset depends on the performance of the classifier \cite{zhou2022feature}.Embedded methods integrate feature weights learning into the training process of the model. After model training, feature selection can be performed based on the learned feature weights. Typical methods for embedded feature selection include sparse regularization-based methods \cite{yang2011,nie2010efficient,l21} and K-nearest neighbor-based methods \cite{bugata2019weighted}. Compared to filter and wrapper methods, embedded methods typically achieve better performance as they consider nonlinear interactions between features. In addition, it can be noted that in recent years, feature selection methods based on deep learning have received widespread attention, especially in the context of extended methods for neural networks \cite{lee2022self,lemhadri2021lassonet}.

\section{The proposed method}\label{sec3}
\subsection{Notations and definitions}\label{subsec2}
Given a matrix ${\bf{M}}\in{{\bf{R}}^{{n \times m}}}$,we denote its ($i$,$j$) entry,$i$-th row,$j$-th column as ${{\bf{M}}_{ij}}$,${{\bf{M}}_i}$,${{\bf{M}}^j}$ respectively.In this paper,${\bf{X}} = {[{x_1},{x_2}, \cdots ,{x_n}]^T} \in {{\bf{R}}^{n \times m}}$ denotes the matrix of data set with missing values,and ${\bf{y}} = {[{y_1},{y_2}, \cdots ,{y_n}]^T}$ denotes the labels corresponding to the samples.
\subsection{Learning algorithm}\label{subsec3}
For incomplete datasets with redundant and noisy features, traditional imputation methods cannot achieve satisfactory imputation results because they do not consider the importance of features. To address this issue that incorporates feature importance into missing value imputation, we propose a two-stage iterative method. Specifically, at the beginning of training, we set each feature corresponds to a weight of 1, indicating that all features have equal importance. Then, in each iteration, the method performs the following two steps: (1) imputing missing values based on the original incomplete data set and the current feature weighting vector $\bf{w}$, and (2) recalculating the feature weighting vector $\bf{w}$ based on the imputed dataset to guide the next round of imputation. The iteration stops when the convergence criterion is met. In particular, we refer to the first and second steps of the iteration as the M-stage and W-stage, respectively. We provide a detailed description of the method below.
\subsubsection{M-stage}\label{subsec4}
The goal of this stage is to perform missing value imputation based on the original incomplete data set and the given feature weighting vector $\bf{w}$. To take into account the importance of features during imputation, we define the imputation loss function as follows:

\begin{flalign}
{\mathop{\min }\limits_{{\bf{G,H}}} {\rm{ }}\psi ({\bf{G}},{\bf{H}}) = \sum\limits_{(p,q) \in \Omega } {{\bf{w}}_q^2{{({{\bf{G}}_p}{{\bf{H}}^q} - {{\bf{x}}_{pq}})}^2} + }\beta (\left\| {\bf{G}} \right\|_F^2 + \left\| {\bf{H}} \right\|_F^2)}
\end{flalign}

where the matrices ${\bf{G}}\in{{\bf{R}}^{{n \times r}}}$ and ${\bf{H}}\in{{\bf{R}}^{{r \times m}}}$ are both low-rank matrices with rank $r$, and their product ${\bf{G}\bf{H}}$ is a matrix for completing {\bf{X}}.$\Omega  = \{ (p,q)|{{\bf{X}}_{pq}}\text{ is observable}\}$denotes the index set of all observable elements in $\bf{X}$, and $\bf{w}$ is the feature weighting vector. Specifically, the imputation loss function of the incomplete data set can be further transformed into an optimization problem in the following form:

\begin{flalign}
{ {\mathop {\min }\limits_{{\bf{G,H}}} {\rm{ }}\psi ({\bf{G,H}}) = \left\| {{\bf{GH}}{\rm{ Diag}}({\bf{w}}) - {\bf{\hat X}}{\rm{ Diag}}({\bf{w}})} \right\|_F^2}+ \beta (\left\| {\bf{G}} \right\|_F^2 + \left\| {\bf{H}} \right\|_F^2)}
\end{flalign}
Where ${\rm{Diag}}({\bf{w}}) = \left[ {\begin{array}{*{20}{c}}
{{w_1}}&0& \cdots &0\\
0&{{w_2}}& \cdots &0\\
 \cdots & \cdots & \cdots & \cdots \\
0&0& \cdots &{{w_m}}
\end{array}} \right]$,and ${\bf{\hat X}} = {P_\Omega }({\bf{X}}) + {P_{\bar \Omega }}({\bf{GH}})$,
$\bar \Omega  = \{ (p,q)|{{\bf{x}}_{pq}}{\text{ is unobservable}}\}$ is the index set of all missing elements in $\bf{X}$, and the function ${P_\Omega }({\bf{X}})$ is defined as follows:
\begin{equation}
{{[{P_\Omega }({\bf{X}})]_{pq}} = \left\{ {\begin{array}{*{20}{c}}
{{{\bf{x}}_{pq}},{\rm{  }}(p,q) \in \Omega }\\
{0,{\rm{     }}(p,q) \notin \Omega }\end{array}} \right.}
\end{equation}

The optimization problem in Eq.(2) is solved using an alternating iterative method. First, we initialize $\bf{G}$ as a random matrix ${{\bf{G}}^{(0)}}$ with orthonormal columns. Then, at the k-th iteration, we calculate ${{\bf{H}}^{(k)}}$ based on $\bf{G}$ = ${{\bf{G}}^{(k-1)}}$, and then calculate ${{\bf{G}}^{(k)}}$ based on $\bf{H}$ = ${{\bf{H}}^{(k)}}$, until the stopping criterion is reached. When $\bf{G}$ is fixed as ${{\bf{G}}^{(k-1)}}$, the optimization problem in Eq.(2) can be simplified to:

\begin{equation}
{{{\bf{H}}^{(k)}} = \mathop {\arg {\rm{min}}}\limits_{{\bf{H}} \in {{\bf{R}}^{r \times m}}} \sum\limits_{q = 1}^m {{\bf{w}}_q^2\left\| {{{\bf{G}}^{\left( {k - 1} \right)}}{{\bf{H}}^q}{\bf{ - }}{{{\bf{\hat X}}}^q}} \right\|_2^2 + \beta \sum\limits_{q = 1}^m {\left\| {{{\bf{H}}^q}} \right\|_2^2} }}
\end{equation}

Eq.(4) can be decomposed into $m$ independent subproblems:

\begin{equation}
    {\mathop {\arg \min }\limits_{{{\bf{H}}^q} \in {{\bf{R}}^{r \times 1}}} f({{\bf{H}}^q}) = {\bf{w}}_q^2\left\| {{{\bf{G}}^{\left( {k - 1} \right)}}{{\bf{H}}^q}{\bf{ - }}{{{\bf{\hat X}}}^q}} \right\|_2^2+\beta \left\| {{{\bf{H}}^q}} \right\|_2^2{\rm{       }}\quad q = 1,2, \cdots ,m.}
\end{equation}
For the $q$-th sub-problem, setting its derivative with respect to the parameter ${{\bf{H}}^q}$ to 0 yields the closed-form solution:
\begin{equation}
    {{\left( {{{\bf{H}}^{(k)}}} \right)^q} = \frac{{{\bf{w}}_q^2}}{{{\bf{w}}_q^2 + \beta }}{({{\bf{G}}^{(k - 1)}})^T}{{\bf{\hat X}}^q}}
\end{equation}
This means that we can calculate ${{\bf{H}}^{(k)}}$ quickly. When $\bf{H}$ is fixed to ${{\bf{H}}^{(k)}}$, the optimization problem in Eq.( 2 ) can be transformed into the following form:

\begin{equation}
    {{{\bf{G}}^{(k)}} = \mathop {\arg \min }\limits_{{\bf{G}} \in {{\bf{R}}^{n \times r}}} \sum\limits_{p = 1}^n {\left\| {{{\bf{G}}_p}{{\bf{H}}^{(k)}}{\rm{Diag}}({\bf{w}}) - {{{\bf{\hat X}}}_p}{\rm{Diag}}({\bf{w}})} \right\|_2^2} + \beta \sum\limits_{p = 1}^n {\left\| {{{\bf{G}}_p}} \right\|_2^2}}
\end{equation}

Eq.( 7 ) can be decomposed into $n$ independent optimization subproblems :

\begin{flalign}
   {\mathop{\min }\limits_{{\bf{G,H}}} {\rm{ }}\psi ({\bf{G}},{\bf{H}}) = \sum\limits_{(p,q) \in \Omega } {\bf{w}}_q^2{{({{\bf{G}}_p}{{\bf{H}}^q} - {{\bf{x}}_{pq}})}^2} +\beta (\left\| {\bf{G}} \right\|_F^2 + \left\| {\bf{H}} \right\|_F^2) }
\end{flalign}

For the $p$-th subproblem, setting its derivative with respect to the parameter ${{\bf{G}}_p}$ to 0 yields the closed-form solution :

\begin{equation}
{\left( {{\bf{G}^{(k)}}} \right)_p} = {{{\bf{\hat X}}}_p}{({\mathop{\rm Diag}\nolimits} ({\bf{w}}))^2}{\left( {{{\bf{H}}^{(k)}}} \right)^T}\times{\left[ {{{\bf{H}}^{(k)}}{{({\mathop{\rm Diag}\nolimits} ({\bf{w}}))}^2}{{\left( {{{\bf{H}}^{(k)}}} \right)}^T} + \beta {\bf{I}}_r} \right]^{ - 1}}
\end{equation}
where ${{\bf{I}}_r}$ is the $r$ × $r$ identity matrix. Upon convergence, the feature matrix with imputed missing values can be obtained by the product of ${{\bf{G}}^{(k)}}$ and ${{\bf{H}}^{(k)}}$. In this paper, we set the convergence criterion as follows:

\begin{equation}
{\begin{array}{l}
\left| {\left\| {{{\bf{H}}^{(k)}}} \right\|_F^2 - \left\| {{{\bf{H}}^{(k - 1)}}} \right\|_F^2} \right| < \eta \\
\left| {\left\| {{{\bf{G}}^{(k)}}} \right\|_F^2 - \left\| {{{\bf{G}}^{(k - 1)}}} \right\|_F^2} \right| < \eta 
\end{array}}
\end{equation}
The threshold $\eta$ is a small positive number, and $\left\|  \bullet  \right\|_F^2$ represents the Frobenius norm of a matrix.

\subsubsection{W-stage}\label{subsec5}
The purpose of this stage is to learn the weighting vector $\bf{w}$ that reflects the importance of features based on the M-stage imputed dataset. To obtain the weighting vector, existing feature selection algorithms such as RFS\cite{nie2010efficient}, WKNN-FS\cite{bugata2019weighted}, and NCFS\cite{yang2012neighborhood} can be used. Considering that NCFS can handle high-dimensional data well and has been widely used in various fields, we adopt it as the method to learn feature importance. In the following, we briefly introduce NCFS.

NCFS is an embedded feature selection method based on a nearest neighbor model. It first initializes the feature importance vector $\bf{w}$ as a vector of all ones. Then, based on $\bf{w}$, the weighted distance of two samples is defined as:
\begin{equation}
{d({\tilde x_i},{\tilde x_j}) = \sum\limits_{l = 1}^m {{\rm{w}}_l^2\left| {{{\tilde x}_{il}} - {{\tilde x}_{jl}}} \right|}}
\end{equation}

Where $w_l$ represents the weight of the l-th feature. In order to learn $\bf{w}$ based on the approximate leave-one-out classification accuracy, NCFS further defines the probability of selecting sample ${\tilde x_i}$with sample ${\tilde x_j}$ as the reference point:
\begin{equation}
{{p_{ij}} = \frac{{\kappa (d({{\tilde x}_i},{{\tilde x}_j}))}}{{\sum\limits_{j = 1,j \ne i}^n {\kappa (d({{\tilde x}_i},{{\tilde x}_j}))} }}}
\end{equation}

Where $\kappa(x) = exp(-x/\sigma)$ is the kernel function, and $\sigma$ is the kernel width.According to the above definition, the probability that the query point $x_i$ is correctly classified is:
\begin{flalign}
    {{p_i} = \sum\limits_j {{y_{ij}}{p_{ij}}} }
\end{flalign}
where $y_{ij}$ = 1 if and only if $y_i$ = $y_j$ otherwise $y_{ij}$ = 0.Finally, NCFS defines the objective function in the following form:
\begin{equation}
{F({\bf{w}}) = \sum\limits_{i = 1}^n {{p_i}}  - \lambda \sum\limits_{l = 1}^m {{\bf{w}}_l^2} }
\end{equation}

Where $\lambda$ is the regularization parameter that needs to be adjusted. Algorithm 1 presents the pseudocode for the proposed imputation method,we called it iterative weighted matrix completion ( IWMC ). In the M-stage, the matrix is imputed based on the current feature weighting vector $\bf{w}$, and then in the W-stage, the feature weighting vector $\bf{w}$ is learned via NCFS based on the imputed dataset. The two stages are executed alternately until the convergence criterion is met, at which point the iteration stops.
\begin{algorithm}
\floatname{algorithm}{Algorithm}
\renewcommand{\algorithmicrequire}{\textbf{Procedure:}}
\renewcommand{\thealgorithm}{1:}
        \caption{Iterative Weighted Matrix Completion for missing value imputation}
        \begin{algorithmic}[1] %
        
            \Require $\textbf{X}$($n\times m$): training set with missing values, $\textbf{y}$($n\times 1$): labels corresponding to the samples, $r$: rank of $\textbf{G}$ and $\textbf{H}$,   $\beta$: regularization parameter, $\Delta$: small positive constant, $\eta$: convergence threshold for $M$-stage. 
            \\Initialization:  \textbf{$\textbf{w}^{(0)}$}= (1,1,...,1) , {$\zeta^{(0)}=-\infty$} ,{$v=0$.}
            \\\textbf{Repeat}
             \State \quad\textbf{{$M$}-stage:}
            \State\quad\quad{Initialize $ \textbf{G}^{(0)} $as a column orthogonal }
            \algnotext{}{random matrix;}
                        \State\quad\quad$k=0$;
            \State\quad\quad\textbf{Repeat}  
            \State\quad\quad\quad$k = k + 1$;
            \State\quad\quad\quad{{\textbf{for}}} {$q$ = 1 ,..., $m$} {{\textbf{do}}}
            \State\quad\quad\quad\quad{Compute $
{\left( {{{\bf{H}}^{(k)}}} \right)^q}$according to $\textbf{w}^{(v)}$ }
            \algnotext{}{and Eq.(6);}
            \State\quad\quad\quad{{\textbf{for}}} {$p$ = 1 ,..., $n$} {{\textbf{do}}}
            \State\quad\quad\quad\quad{Compute $
{({{\bf{G}}^{(k)}})_p}$ according to $\textbf{w}^{(v)}$ } 
            \algnotext{}{Eq.(9);}
            \State\quad\quad\textbf{Until} { the conditions in Eq.(10) are }
            \algnotext{}{satisfied}
             \State\quad\quad{$\tilde{\textbf{X}}= {P_\Omega }(X) + {P_{\bar \Omega }}({{\bf{G}}^{(k)}}{{\bf{H}}^{(k)}})$};
             \State\quad\textbf{{$W$}-stage:}
             \State\quad\quad $v=v+1$;
             \State\quad\quad Use $NCFS$ to compute $\textbf{w}^{(v)}$ based on $\tilde{\textbf{X}}$  
             \algnotext{}{and $\textbf{y}$;}
            \State\quad\quad {$\zeta^{(v)}= \left\| {{{\bf{\textbf{w}}}^{(v)}}} \right\|_2^2$};
           \\ \textbf{Until}  {$|\zeta^{(v)}-\zeta^{(v-1)}|<\Delta$}
           \State \textbf{Return} $\textbf{w}^{(v)}$ and  $\tilde{\textbf{X}}$ 
           \end{algorithmic}
    \end{algorithm}

\begin{table*}
\renewcommand\arraystretch{1.3}
  \caption{Datasets}\centering
  \begin{tabular*}{0.81\linewidth}{lllllll}
  \hline
     No.&	 Dataset& Type/Source	& Instances& Features& Classes&\thead{ Missing\\  rate(\%)}\\
     \hline
     1 &  lung\_discrete& \multirow{3}{*}{Biology} & 73& 325&7 &{0}\\
     2 &  colon&  & 62 &2000  & 2 &{0}\\
    3 & lymphoma&  & 96 & 4026 & 9 &{0}\\
     4 & warpPIE10P&\multirow{2}{*}{Image}  & 210		 & 2420 & 10 &{0}\\
     5 & warpAR10P& 		 &130  & 2400 & 10 &{0}\\
     \hline
    6 & Autism Screening& \multirow{2}{*}{UCI}	& 704	& 20	& 2	& 1.36\\

    7 & Autistic Spectrum	&	&292	&20	&2	&1.54\\  
     8 & ASP-POTASSCO&OpenML & 1294		 & 141 & 11&10\\
     9 &HCC survival	& \multirow{2}{*}{UCI}	& 165	& 49	& 2	& 10.21
\\
    10 & cervical cancer	& 	& 858	& 35	& 2	& 12.06
\\
     11 & pbc          &OpenML         & 418			 & 18 & 2 &16.46\\
     12 & horse&UCI & 300			 & 27 & 2 &19.81\\
     13 & colic&  \multirow{2}{*}{OpenML} &368			 & 22 & 2 &23.8\\
    14 & AotuMLSelector& 			 & 103 & 108 &  8 &34.17\\    
  \hline
\end{tabular*}
\end{table*}

\section{Experiment}\label{sec4}

In this section, we conducted multiple experiments to evaluate the performance of the proposed algorithm. Firstly, we investigated the parameter sensitivity of the algorithm to determine the optimal parameter settings. Next, we compared the proposed method with other methods on synthetic datasets and real-world datasets, and the experimental results confirmed the effectiveness of the proposed method. Finally, we discussed the stability of the proposed algorithm.
\subsection{Datasets}\label{sec1}
We used fourteen real-world datasets to validate the performance of IWMC, where the first five datasets are facial image datasets and biology datasets\cite{li2017feature}, while the remaining nine datasets are from the UCI\cite{asuncion2007uci} and OpenML\cite{vanschoren2014openml} databases. Table 1 provides detailed information about the fourteen datasets. To conduct missing value experiments on the first five datasets, we generated missing values at three levels (1\%, 5\%, and 10\%) based on two missing mechanisms:missing completely at random(MCAR) and missing not at random(MNAR). The remaining nine datasets have different degrees of missing values, which do not require artificial missing value generation.In addition, we created an synthetic dataset using the synthetic dataset generation method provided by scikit-learn\cite{pedregosa2011scikit}. This dataset was used in the feature selection challenge of NIPS 2003\cite{guyon2003introduction} and was a binary classification dataset,we set it 300 samples and 10 features initially without missing values. We added different noise features and missing values to it for different experiments. Finally, we standardized all datasets to have a normal distribution.

\subsection{Parameter sensitivity analysis}\label{sec2}
To evaluate the sensitivity of the proposed method to parameter settings, we investigated how the performance of the IWMC algorithm is affected by the values of the M-phase parameters $r$ and $\beta$. As the parameters $\lambda$ and $\sigma$ for NCFS have been studied in \cite{yang2012neighborhood} and are recommended to be set to \{1, 1\}, we focused only on the parameters $r$ and $\beta$ in this paper. We searched for the optimal settings of parameter $r$ from \{1, 3, 5, 10\} and parameter $\beta$ from \{0.25, 0.5, 1, 2, 5, 10, 20, 30\} that yield the best performance. We generated 6 different datasets for the parameter sensitivity experiments by adding noise features to the synthetic datasets discribed in 4.1 and created 5\% and 20\% MCAR data. The noise features were extracted from random numbers with a mean of 0 and a variance of 5. As the 10 relevant features in the synthetic dataset are known in advance, we ranked the feature weights learned by our proposed method on each dataset and selected the top 10 features with the highest weights. We then counted the number of relevant features among these 10 features. We used the success rate of the algorithm in selecting relevant features and the corresponding standard deviation of the success rate over 10 runs as the evaluation metric. The standard deviation provides an intuitive measure of the stability of the results with respect to parameter settings.The experimental results with different parameters are shown in Figures 1-4.

Based on the results shown in Figures 1-4, we made the following observations: (1) The parameter $r$ has little influence on the performance and stability of the proposed algorithm. (2) In most cases, the highest success rate for feature selection was achieved when $\beta$ was set to 10 or 20. (3) The proposed algorithm exhibited the best stability when $\beta$ was set to 20 or 30. Therefore, we recommend setting $\beta$ to 20 and $r$ to 5.

\begin{figure*}[t]
  \centering
  \includegraphics[width=\linewidth]{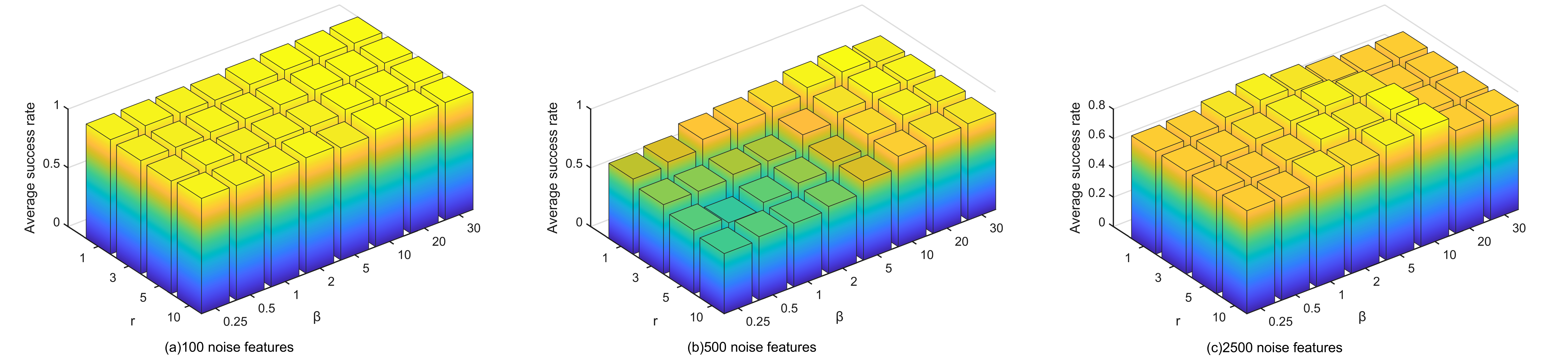}
  \caption{Success rate of relevant features selected on the synthetic dataset with 5\% of missing data.}\centering
\end{figure*}

\begin{figure*}[t]
  \centering
  \includegraphics[width=\linewidth]{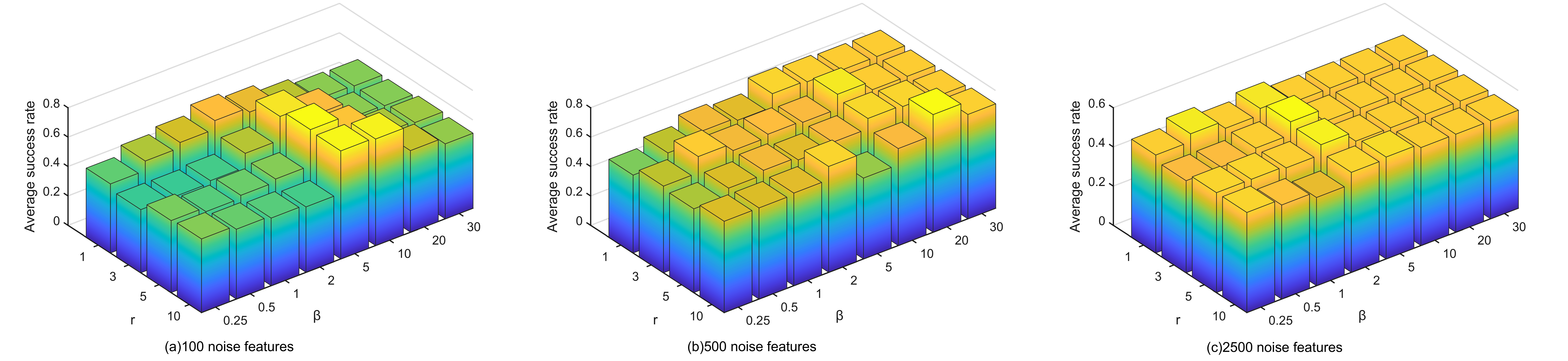}
  \caption{Success rate of relevant features selected on the synthetic dataset with 20\% of missing data.}\centering
\end{figure*}

\begin{figure*}[t]
  \centering
  \includegraphics[width=\linewidth]{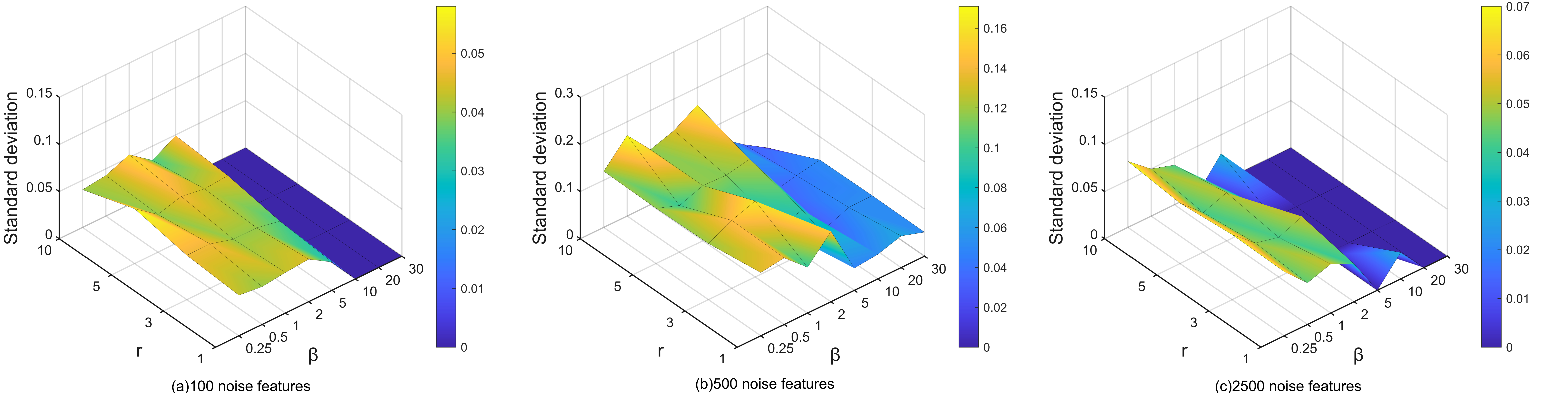}
  \caption{The standard deviation of the success rate of relevant feature selected on the synthetic dataset with 5\% of missing data.}\centering
\end{figure*}
\begin{figure*}[t]
  \centering
  \includegraphics[width=\linewidth]{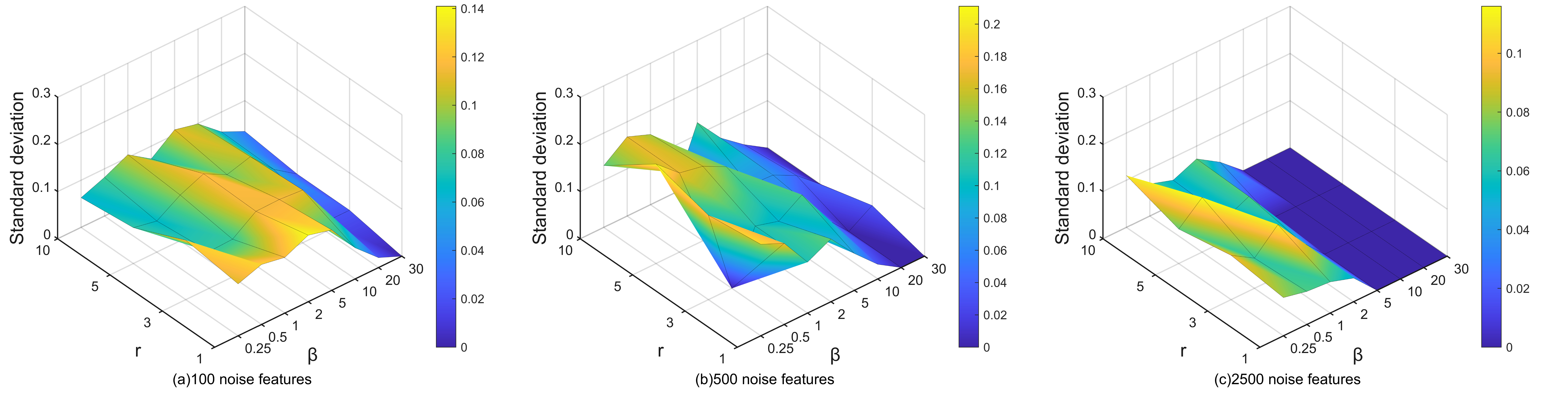}
  \caption{The standard deviation of the success rate of relevant feature selected on the synthetic dataset with 20\% of missing data.}\centering
\end{figure*}

\begin{table*}
\renewcommand\arraystretch{1.5}
  \caption{${F_{1}}$-score of 5NN using 5-fold cross validation on datasets with MCAR data(Mean±std).}\centering
  \label{tab:commands}
  \begin{tabular*}{0.86\linewidth}{l|l|llllll}
  \hline

{\small Data} &
  {\small Missing} &
\multirow{2}{*}{{\small IWMC}}  &
  {\small Mean} &
  {\small fast KNN} &
  {\small EM} &
  {\small I-SVD} &
  {\small SOFT} \\ \cmidrule{4-8} 
                         {\small sets}   & {\small rate}     &         & \multicolumn{5}{c}{{\small{Then use NCFS to select the top 50 features}}}                          \\
 
    \hline
                   \multirow{3}*{\rotatebox{90}{\small lungdiscrete} } & \makecell[c]{1\%}   & \small\bfseries{81.36±2.52}   &\small77.28±3.64 &	\small79.21±2.45&	\small78.38±2.98	& \small77.07±4.41 &	\small78.01±4.63      \\
    & \makecell[c]{5\%}   & \small\textbf{81.06±2.28}   &\small77.14±3.70 & \small77.17±2.72 & \small78.31±4.41 & \small76.04±3.12	& \small77.95±3.84 \\
                    & \makecell[c]{10\%}  & \small\textbf{79.61±2.07}  & \small76.48±3.25 &	\small76.66±4.18	& \small74.79±4.25	& \small74.96±4.52 &	\small76.36±4.32\\
\cmidrule{1-2}
           \multirow{3}*{\rotatebox{90}{\small colon} }         & \makecell[c]{1\%}   & \small\textbf{82.88±2.51}&	\small80.07±4.96&	\small82.11±4.55	&\small79.21±4.02	&\small80.92±1.85	&\small79.45±3.56    \\
    & \makecell[c]{5\%}   & \small\textbf{82.68±1.97}	&\small78.84±3.63	&\small79.32±3.34 &\small78.90±5.62	&\small80.40±6.11&	\small78.81±4.29    \\
                    & \makecell[c]{10\%}  & \small\textbf{81.02±2.58}	&\small76.23±5.12	&\small78.06±4.76	&\small77.61±2.53	&\small80.33±3.56	&\small78.71±6.15\\
\cmidrule{1-2}
\multirow{3}*{\rotatebox{90}{\small \small warpAR10P} }& \makecell[c]{1\%}   & \small\textbf{82.88±2.51}&	\small80.07±4.96&	\small82.11±4.55	                                &\small79.21±4.02	&\small80.92±1.85	&\small79.45±3.56    \\
    & \makecell[c]{5\%}   & \small74.15±2.31&	\small\textbf{77.14±3.82}	&\small75.07±2.83	&\small75.62±4.22	&\small75.02±5.10&	\small76.34±2.22        \\
                    & \makecell[c]{10\%}  & \small63.92±3.11&	\small74.45±4.49&	\small73.85±5.06&	\small69.58±5.01&	\small\textbf{74.97±5.44}&	\small73.82±3.44\\
\cmidrule{1-2}
\multirow{3}*{\rotatebox{90}{\small \small warpPIE10P} }& \makecell[c]{1\%}   & \small\textbf{98.04±1.16}&	\small96.73±1.57&	\small97.17±0.66	&\small97.11±0.76&	\small96.78±0.88&	\small97.33±1.15   \\
    & \makecell[c]{5\%}   & \small96.94±1.34&	\small96.68±0.86&	                                   \small\textbf{97.11±1.11}&	\small96.65±1.95&	\small96.41±1.27	&\small97.04±1.41 \\
                    & \makecell[c]{10\%}  & \small95.04±0.98	&\small96.07±1.45&	\small96.25±1.08&	\small95.58±1.66&	\small95.63±1.35	&\small\textbf{96.81±1.16}\\
\cmidrule{1-2}
       \multirow{3}*{\rotatebox{90}{\small \small lymphoma} } & \makecell[c]{1\%}    & \small\textbf{89.89±0.72}	& \small82.35±4.74	& \small81.60±2.53	& \small84.16±2.92	& \small83.15±3.46	& \small83.16±3.85\\
    & \makecell[c]{5\%}   & \small\textbf{88.74±0.21}	& \small81.81±3.90	& \small80.7±3.97	& \small80.95±4.41	& \small82.97±4.13	& \small81.80±3.18\\
                    & \makecell[c]{10\%}  & \small\textbf{85.75±1.94}	&\small80.46±3.62	&\small80.19±3.45	&\small79.56±4.09	&\small82.62±3.02	&\small79.77±3.67\\            

    \hline
  \end{tabular*}
\end{table*}

\begin{table*}
\renewcommand\arraystretch{1.5}
  \caption{ACC of 5NN using 5-fold cross validation on datasets with MCAR data(Mean±std).}\centering
  \label{tab:commands}
  \begin{tabular*}{0.86\linewidth}{l|l|llllll}
  \hline
 
{\small Data} &
  {\small Missing} &
\multirow{2}{*}{{\small IWMC}}  &
  {\small Mean} &
  {\small fast KNN} &
  {\small EM} &
  {\small I-SVD} &
  {\small SOFT} \\ \cmidrule{4-8} 
                         {\small sets}   & {\small rate}     &         & \multicolumn{5}{c}{{\small{Then use NCFS to select the top 50 features}}}                           \\
    \hline

                \multirow{3}*{\rotatebox{90}{\small lungdiscrete} }   & \makecell[c]{1\%}   & \small\bfseries{82.43±2.17}   &\small77.45±3.36 &	\small80.16±2.77&	\small78.56±2.66	& \small77.55±4.11 &	\small79.19±3.08      \\
   & \makecell[c]{5\%}   & \small\textbf{81.12±2.16}   	&\small77.37±3.57	&\small78.67±3.73&	\small77.73±5.55&	\small77.49±2.83&	\small78.27±3.91 \\
                    & \makecell[c]{10\%}  & \small\textbf{79.31±2.84}  & 	\small77.16±2.94	&\small78.01±2.76	&\small76.32±4.11&	\small76.16±3.53&	\small77.35±3.68\\
\cmidrule{1-2}
                    \multirow{3}*{\rotatebox{90}{\small colon}}& \makecell[c]{1\%}   & \small\textbf{83.25±1.99}&		\small79.87±4.86	&\small81.88±4.92&	\small79.09±4.24&	\small80.89±1.52	&79.21±4.09 \\
    & \makecell[c]{5\%}   & \small\textbf{83.16±1.76}&		\small78.70±3.77	&\small79.24±3.37&	\small78.87±4.99&	\small80.62±3.39&	\small78.80±6.21    \\
                    & \makecell[c]{10\%}  & \small\textbf{81.67±2.15}	&	\small76.10±5.02&	\small78.11±4.45&	\small78.26±2.77	&\small80.22±5.78&	\small78.69±4.47\\
\cmidrule{1-2}
\multirow{3}*{\rotatebox{90}{\small warpAR10P}}& \makecell[c]{1\%}   & \small\textbf{80.20±2.10}&		\small77.59±2.92	&\small77.78±3.24	&\small76.44±4.06&	\small75.01±2.21&	\small77.84±5.04    \\
   {\small } & \makecell[c]{5\%}   & \small74.15±1.42	&\small\textbf{76.15±4.25}&	\small74.61±2.85&	\small75.69±3.77&	\small75.43±4.32&	\small75.69±2.35      \\
                    & \makecell[c]{10\%}  & \small65.62±2.23&	\small73.84±4.02&	\small73.17±4.53&	\small69.42±4.71&	\small\textbf74.46±5.71&	\small73.38±3.78\\
\cmidrule{1-2}
\multirow{3}*{\rotatebox{90}{\small warpPIE10P}}& \makecell[c]{1\%}   & \small\textbf{98.04±1.23}	& 	\small96.66±1.58	&\small97.10±0.68	&\small96.99±0.82&	\small96.52±0.89&	\small97.23±1.22\\
   & \makecell[c]{5\%}   & \small\textbf{97.24±1.16}&\small		96.63±0.85	&\small97.08±1.06	&\small96.66±1.83&\small	96.28±1.34&\small	96.95±1.45\\
                    & \makecell[c]{10\%}  &\small 95.32±0.98&\small	95.95±1.42	&\small96.14±1.17	&\small95.52±1.69	&\small95.47±1.41	&\small\textbf{96.76±1.22}\\
\cmidrule{1-2}
       \multirow{3}*{\rotatebox{90}{\small lymphoma}} & \makecell[c]{1\%}    & \small\textbf{91.76±0.63}	& \small85.53±3.71	& \small84.33±2.21	& \small86.76±2.42	& \small85.73±3.56	& \small85.80±3.02\\
   & \makecell[c]{5\%}   & \small\textbf{89.86±0.92}	& \small84.14±3.76	& \small83.57±3.47	& \small83.38±4.09	& \small85.41±2.64	& \small84.51±2.54\\
                    & \makecell[c]{10\%}  & \small\textbf{87.48±1.26}	& \small83.61±3.51	& \small83.11±2.77	& \small83.11±3.71	& \small84.97±2.79	& \small82.73±3.73\\            
    \hline
  \end{tabular*}
\end{table*}

%
%
\begin{table*}
\renewcommand\arraystretch{1.5}
  \caption{${F_{1}}$-score of 5NN using 5-fold cross validation on datasets with MNAR data(Mean±std).}\centering
  \label{tab:commands}
  \begin{tabular*}{0.86\linewidth}{l|l|llllll}
  \hline
 
{\small Data} &
  {\small Missing} &
\multirow{2}{*}{{\small IWMC}}  &
  {\small Mean} &
  {\small fast KNN} &
  {\small EM} &
  {\small I-SVD} &
  {\small SOFT} \\ \cmidrule{4-8} 
                         {\small sets}   & {\small rate}     &         & \multicolumn{5}{c}{{\small{Then use NCFS to select the top 50 features}}}                          \\

    \hline
               \multirow{3}*{\rotatebox{90}{\small lungdiscrete} }     & \makecell[c]{1\%}   & \small\bfseries{80.73±3.32}	& \small78.93±5.41	& \small75.86±5.61	& \small79.29±5.41	& \small78.21±4	& \small78.38±4.19\\
    & \makecell[c]{5\%}   & \small{74.87±3.17}	& \small74.82±3.62	& \small6.42±2.82	& \small74.04±3.447	& \small74.79±4.37	& \small\textbf{77.69±3.05} \\
                    & \makecell[c]{10\%}  & \small52.73±6.69  & \small60.75±4.92 	& \small\textbf{70.55±3.02}	& \small51.14±3.94	& \small58.34±5.84	& \small58.91±5.58
\\
\cmidrule{1-2}
           \multirow{3}*{\rotatebox{90}{\small colon} }         & \makecell[c]{1\%}   & \small{81.18±2.72}	& \small80.36±3.8	& \small\textbf{82.6±3.3}	& \small81.71±3.16	& \small79.32±3.06	& \small78.99±6.64   \\
    & \makecell[c]{5\%}   & \small\textbf{78.64±2.85}	& \small71.92±3.59	& \small75.21±4.73	& \small77.81±2.56	& \small76.52±4.12	& \small74.62±2.81    \\
                    & \makecell[c]{10\%}  & \small\textbf{73.17±2.7}	& \small71.53±3.16	& \small72.96±4.09	& \small72.31±3.65	& \small69.43±5.79	& \small72.5±5.19\\
\cmidrule{1-2}
\multirow{3}*{\rotatebox{90}{\small warpAR10P} }& \makecell[c]{1\%}   & \small\textbf{76.85±2.77}& \small76.7±4.37	& \small76.23±2.62	& \small76.11±4.37	& \small76.42±2.04	& \small76.67±3.52
  \\
 & \makecell[c]{5\%}   & \small\textbf{72.99±2.86}	& \small70.63±3.77	& \small70.21±3.95	& \small69.15±2.81	& \small72.65±4.28	& \small72.93±3.55
        \\
                    & \makecell[c]{10\%}  & \small69.45±3.37	& \small64.55±3.78	& \small70.67±2.8	& \small65.07±4	& \small\textbf{70.98±3.3}	& \small70.73±4.36
\\
\cmidrule{1-2}
\multirow{3}*{\rotatebox{90}{\small warpPIE10P} }& \makecell[c]{1\%}   & \small\textbf{96.66±1.22}	& \small96.14±1.61	& \small96.49±1.58	& \small96.43±1.07	& \small96.49±1.58	& \small96.4±1.33 \\
    & \makecell[c]{5\%}   & \small92.58±1.68	& \small95.67±0.94	& \small95.2±1.16	& \small93.52±1.48	& \small96.25±1.08	& \small\textbf{96.28±1.04} \\
                    & \makecell[c]{10\%}  & \small83.84±1.91	& \small93.48±1.57	& \small93.39±1.48	& \small89.89±1.37	& \small95.67±0.99	& \small96.13±0.76\\
\cmidrule{1-2}
    \multirow{3}*{\rotatebox{90}{\small lymphoma} }    & \makecell[c]{1\%}    & \small\textbf{85.84±2.57}	& \small80.93±2.84	& \small83.6±2.99	& \small83.28±3.08	& \small81.61±2.69	& \small83.48±2.98\\
    & \makecell[c]{5\%}   & \small\textbf{81.61±2.77}	& \small78.33±3.61	& \small81.51±2.72	& \small80.81±4.04	& \small80.31±2.77	& \small77.19±3.78\\
                    & \makecell[c]{10\%}  & \small\textbf{75.5±2.86}		& \small69.38±4.23		& \small74.7±3.89		& \small74.46±3.31		& \small73.39±5.74		& \small72.9±6.12\\            
    \hline
  \end{tabular*}
\end{table*}

\begin{table*}
\renewcommand\arraystretch{1.5}
  \caption{ACC of 5NN using 5-fold cross validation on datasets with MNAR data(Mean±std).}\centering
  \label{tab:commands}
  \begin{tabular*}{0.86\linewidth}{l|l|llllll}
  \hline

{\small Data} &
  {\small Missing} &
\multirow{2}{*}{{\small IWMC}}  &
  {\small Mean} &
  {\small fast KNN} &
  {\small EM} &
  {\small I-SVD} &
  {\small SOFT} \\ \cmidrule{4-8} 
                         {\small sets}   & {\small rate}     &         & \multicolumn{5}{c}{{\small{Then use NCFS to select the top 50 features}}}                           \\
    \hline

            \multirow{3}*{\rotatebox{90}{\small lungdiscrete} }        & \makecell[c]{1\%}   & \small\bfseries{81.76±2.94}	& \small78.97±3.67	& \small77.07±5.07	& \small79.43±5.67	& \small78.35±4.31	& \small78.79±3.71\\
    & \makecell[c]{5\%}   & \small76.67±2.78	& \small75.07±2.84	& \small\textbf{77.03±2.58}	& \small74.41±3.58	& \small74.58±3.96	& \small77.5±3.21\\
                    & \makecell[c]{10\%}  & \small56.08±5.96  & \small54.91±3.28	& \small62.76±4.84	& \small\bfseries{71.28±3.86}	& \small60.49±5.66	& \small61.67±5.5
\\
\cmidrule{1-2}
     \multirow{3}*{\rotatebox{90}{\small colon} }               & \makecell[c]{1\%}   & \small\textbf{82.28±2.59}	& \small80.39±3.89		& \small80.56±3.22	& \small79.51±2.94& \small78.83±3.89	& \small79.32±6.01 \\
    & \makecell[c]{5\%}   & \small\textbf{78.87±2.85}	& \small76.22±3.45	& \small73.56±3.05	& \small74.98±4.76	& \small76.44±3.17	& \small75.25±2.56\\
                    & \makecell[c]{10\%}  & \small71.88±2.53		& \small72.89±3.33	& \small73.98±4.64& \small\textbf{74.11±3.37}	& \small71.12±5.09	& \small73.5±5.28
\\
\cmidrule{1-2}
\multirow{3}*{\rotatebox{90}{\small warpAR10P} }& \makecell[c]{1\%}   & \small76.84±2.66	& \small75.84±4.21	& \small76.15±2.78	& \small75±4.63	& \small76±3.14	& \small\textbf{77.38±3.5}
\\
    & \makecell[c]{5\%}   & \small\textbf{73.3±2.65}	& \small71.23±3.8	& \small70.85±2.65	& \small69±3.31	& \small71.84±4.54	& \small73.15±3.31
 \\
                    & \makecell[c]{10\%}  & \small\textbf{71.08±2.82}	& \small64.15±3.55	& \small70±3.05	& \small64.23±3.64	& \small70.62±3.79	& \small70.16±4.3
\\
\cmidrule{1-2}
\multirow{3}*{\rotatebox{90}{\small warpPIE10P} }& \makecell[c]{1\%}   & \small\textbf{96.57±1.24}	& \small96.47±1.23	& \small96.47±1.31	& \small96±1.78	& \small96.31±1.64	& \small96.4±1.23
\\
    & \makecell[c]{5\%}   & \small92.58±1.68	& \small94.9±0.75	& \small\textbf{94.95±1.23}	&\small93.23±1.02	& \small96.07±1.19	& \small96.09±0.99\\
                    & \makecell[c]{10\%}  &\small 84.4±1.84	&\small 93.42±1.45	&\small 93.38±1.41	&\small 89.71±1.42	&\small 95.53±1.02	&\small\textbf{ 95.75±0.99}\\
\cmidrule{1-2}
       \multirow{3}*{\rotatebox{90}{\small lymphoma} } & \makecell[c]{1\%}    & \small\textbf{88.29±2.29}	& \small86.58±2.82	& \small85.69±3.31	& \small83.98±2.86	& \small84.77±2.73	& \small85.85±2.61
\\
    & \makecell[c]{5\%}   & \small\textbf{85.11±2.08}		& \small84.9±2.84 &\small83.77±3.89	& \small82.47±3.2	& \small84.05±3.01& \small81.33±3.18\\
                    & \makecell[c]{10\%}  & \small\textbf{80.86±2.98}	& \small75.31±3.40	& \small80.4±4.99	& \small79.01±2.14	& \small78.9±4.46	& \small78.53±4.88
\\            
    
    \hline
  \end{tabular*}
\end{table*}

\begin{figure*}[h]
  \centering
  \includegraphics[width=\linewidth]{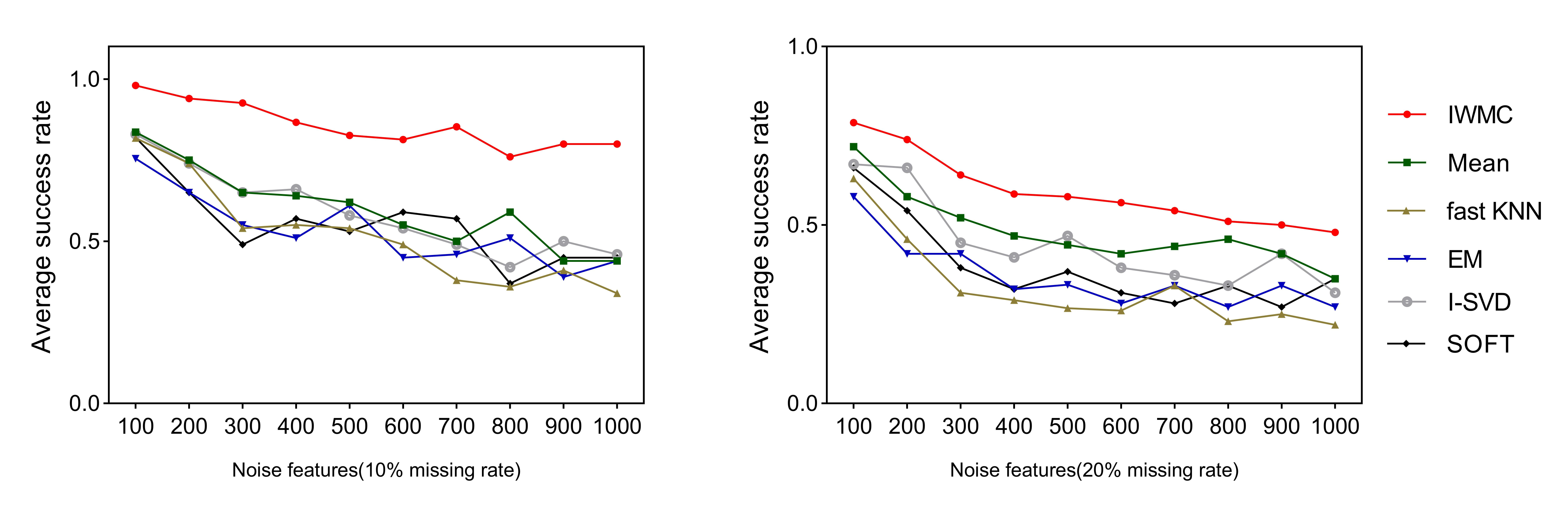}
  \caption{Average success rate of relevant features selected on synthetic dataset with MCAR data.}\centering
\end{figure*}

\begin{figure*}[h]
  \centering
  \includegraphics[width=\linewidth]{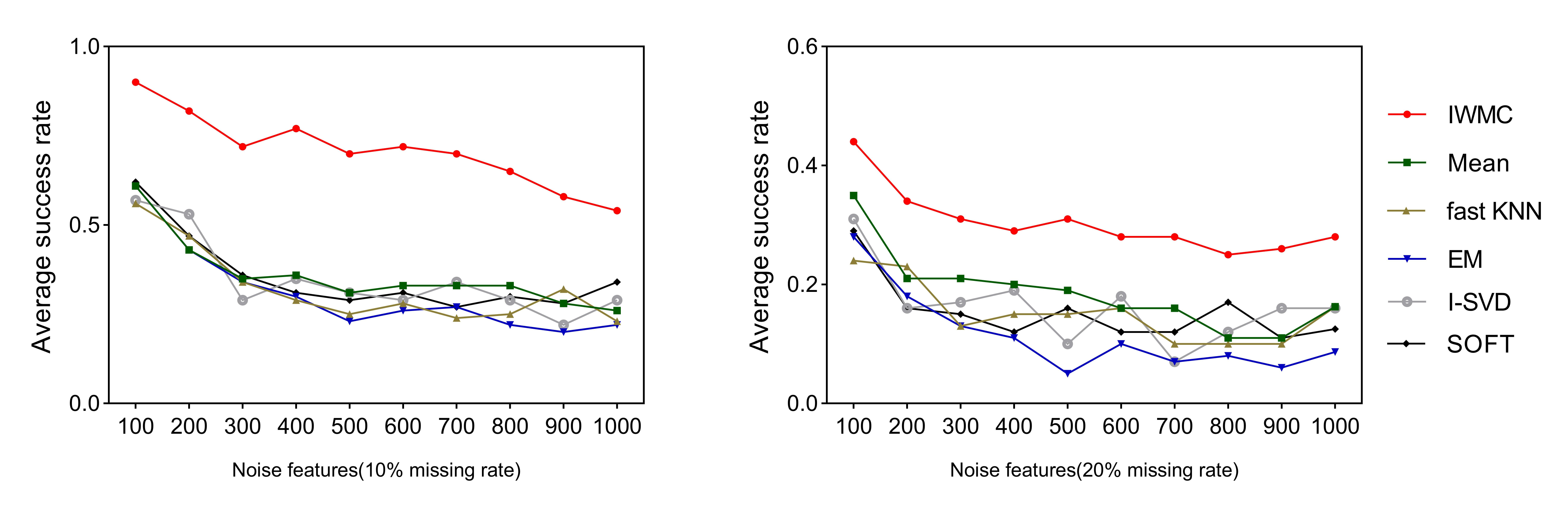}
  \caption{Average success rate of relevant features selected on synthetic dataset with MNAR data.}\centering
\end{figure*}

\subsection{Comparison with other methods}\label{sec3}
To evaluate the proposed algorithm, we compared it with five imputation methods: mean imputation, expectation-maximization (EM) imputation, fast k-nearest neighbor (KNN) imputation, iterative singular value decomposition (I-SVD)\cite{troyanskaya2001missing}, and SOFT imputation\cite{mazumder2010spectral}. \textbf{Mean} imputation imputes the missing values with the mean value of all observed values in the current column. \textbf{EM} imputation is a maximum likelihood estimation method. \textbf{Fast KNN} imputation is an imputation algorithm based on fast KNN. \textbf{I-SVD} approximates the matrix ${{\bf{A}}^{n \times m}}$ by decomposing it into three matrices ${{\bf{U}}^{n \times u}}$, ${{\bf{P}} ^{{{u}} \times {{u}}}}$, and ${{\bf{V}}^{u \times m}}$ multiplied together, using the top $u$ ($u \ll min(n,m)$) largest singular values to approximate the matrix. \textbf{SOFT} imputation minimizes the error function subject to a nuclear norm constraint at each iteration, and then updates the incomplete matrix with the result of soft threshold singular value decomposition (SVD). The algorithm stops when a certain condition is met.

In the experiment, we first used these 5 methods to impute the missing values, and then used NCFS to learn feature importance on the imputed dataset. We evaluated the performance of each imputation method by the success rate of identifying relevant features on the synthetic dataset and the classification performance on the real-world datasets.

\subsection{Results on the synthetic datasets}\label{sec4}
Based on the synthetic datasets with different numbers of noise features \{100, 200, 300, 400, 500, 600, 700, 800, 900, 1000\}, we further compared our proposed method with five algorithms under two missing ratios {10\%, 20\%} and two missing mechanisms MCAR and MNAR. For the comparative methods, we calculated the feature weights using NCFS based on their imputed datasets, and then compared the learned feature weights with our method. Note that ten relevant features are known in advance, so we can select the ten features with the largest weights to calculate the success rate of feature selection. Figures 5 and 6 show the experimental results for MCAR and MNAR, respectively. It can be seen from the figures that the proposed method outperforms the other five methods under both missing mechanisms, and MCAR results are generally better than MNAR results. As the number of noise features increases, the results of IWMC decline more slowly, and each result is consistently higher than the other comparative methods. Another finding is that although mean imputation is the simplest statistical imputation model, as it uses a fixed value for imputation and can only be considered as a rough estimate, the results of mean imputation demonstrate that it is a strong competitor in this task, and its result curve of decline is smoother than that of other comparative methods. This is because when there are not too many outliers in the dataset, the mean can represent the most common information in the data. Under the interference of noise features, the comparative methods imputed data deviate more on the relevant feature items, and cannot iteratively obtain the importance of features, so their performance is worse than our method. Our initial hypothesis was that considering feature weights in missing value imputation can improve the imputation quality of relevant features and help feature selection methods identify relevant features, and the subsequent result analysis also proved this hypothesis.

\subsection{Experiments on real-world datasets}\label{sec5}
In this section, we will investigate the performance of the proposed method on real-world datasets. In the experiment, we choose KNN classifier to calculate the classification accuracy and ${F_{1}}$-score.

\textbf{(1)Accuracy:}
\begin{equation}
    {AC{C_{knn}}({\bf{X}},{\bf{y}}) = \frac{1}{n}\sum\limits_{i = 1}^n {{\rm I}({y_i} = c({x_i}))}}
\end{equation}

Where I$({y_i} = c({x_i}))=1$ if and only if ${y_i} = c({x_i})$, $y_i$ is the true label of $x_i$, and $c(x_i)$ is the predicted label of sample $x_i$ calculated by KNN classifier.

\textbf{(2)$F_{1}$-score}:

In binary classification problems, samples are divided into positive and negative categories. Based on the actual and predicted labels of samples, samples can be classified into four categories: True Positive (TP), False Positive (FP), True Negative (TN), and False Negative (FN). TP refers to the number of positive samples that are correctly classified as positive, FP refers to the number of negative samples that are incorrectly classified as positive, TN refers to the number of negative samples that are correctly classified as negative, and FN refers to the number of positive samples that are incorrectly classified as negative.

Precision refers to the proportion of positive samples that are correctly classified out of all samples that are predicted to be positive, and recall refers to the proportion of positive samples that are correctly classified out of all samples that are actually positive. The definitions of precision and recall are given in Eq.(15) and Eq.(16):
\begin{equation}
    {precision = \frac{{TP}}{{TP + FP}}}
\end{equation}

\begin{equation}
    {recall = \frac{{TP}}{{TP + FN}}}
\end{equation}

Generally, we want to balance both precision and recall. Therefore, ${F_{1}}$-Score is another commonly used metric, which is the harmonic mean of precision and recall and can comprehensively evaluate the performance of the model. Its definition is given by equation (17):
\begin{equation}
    {{F_{1}} = 2 \cdot \frac{{precision \cdot recall}}{{precision + recall}}}
\end{equation}

The $F_{1}$-score for binary classification can be extended to multi-class problems. In multi-class problems, one class is considered as the positive class and the others are considered as negative classes, and then the $F_{1}$-score is calculated according to Eq.(17).

\begin{table*}
\renewcommand\arraystretch{1.5}
  \caption{${F_{1}}$-score of 5NN using 5-fold cross validation on real missing datasets(Mean±std).}\centering
  \label{tab:commands}
  \begin{tabular*}{0.86\linewidth}{l|llllll}
  \hline
    \toprule
{\small Data} &
\multirow{2}{*}{{\small IWMC}}  &
  {\small Mean} &
  {\small fast KNN} &
  {\small EM} &
  {\small I-SVD} &
  {\small SOFT} \\ \cmidrule{3-7} 
                         {\small sets}        &         & \multicolumn{5}{c}{{\small{Then use NCFS to select the top 50\% of the original features}}}                          \\

    \hline
     {\small Autism Screening}              & \small\bfseries{95.11±0.6}	& \small92.8±1.04	& \small91.85±1.35	& \small92.15±0.62	& \small92.7±0.69	& \small92.95±1.02\\
\cmidrule{1-1}
 {\small Autistic Spectrum}              & \small\bfseries{91.85±2.68}	& \small87.17±2.58	& \small86.76±2.36	& \small87.24±1.89	& \small87.56±1.08	& \small86.79±1.65\\
\cmidrule{1-1}
{\small ASP-POTASSC}              & \small\bfseries{39.45±0.83}	& \small39.03±1.01	& \small38.50±1.03	& \small36.66±0.64	& \small38.58±1.36	& \small39.35±0.78\\
\cmidrule{1-1}
{\small HCC survival}              & \small66.02±3.72	& \small61.56±3.41	& \small63.05±2.81	& \small63.7±3.86	& \small66.15±3.83	& \small\bfseries{66.07±3.6}\\
\cmidrule{1-1}
 {\small cervical cancer}              & \small94.13±0.3	& \small94.05±0.59	& \small94.12±0.48	& \small93.99±0.41	& \small\bfseries{94.48±0.53}	& \small94.16±0.39
\\
\cmidrule{1-1}
   {\small pbc}  & \small\textbf{64.39±1.35}	& \small63.96±1.74	& \small64.27±2.25	& \small63.17±1.44	& \small63.67±1.83	& \small64.18±1.43    \\             
\cmidrule{1-1}
   {\small horse}& \small\textbf{79.22±1.62}	& \small69.82±1.93	& \small65.63±1.82	& \small66.04±2.52	& \small70.06±2.03	& \small76.73±1.33\\             
\cmidrule{1-1}

   {\small colic}   & \small\textbf{83.02±1.21}	& \small82.81±1.28	& \small82.55±1.31	& \small82.22±0.67	& \small82.91±1.28	& \small81.82±0.76\\                 
\cmidrule{1-1}   
   {\small AotuMLSelector}  & \small\textbf{48.11±2.27}	& \small40.46±3.45	& \small38.63±2.96	& \small35.85±2.74	& \small39.21±4.76	& \small42.41±1.92\\                    

    \hline
  \end{tabular*}
\end{table*}

\begin{table*}
\renewcommand\arraystretch{1.5}
  \caption{ACC of 5NN using 5-fold cross validation on real missing datasets(Mean±std).}\centering
  \label{tab:commands}
  \begin{tabular*}{0.86\linewidth}{l|llllll}
  \hline

{\small Data} &
\multirow{2}{*}{{\small IWMC}}  &
  {\small Mean} &
  {\small fast KNN} &
  {\small EM} &
  {\small I-SVD} &
  {\small SOFT} \\ \cmidrule{3-7} 
                         {\small sets}        &         & \multicolumn{5}{c}{\small{Then use NCFS to select the top 50\% of the original features}}                          \\

    \hline
    {\small Autism Screening}              & \small\bfseries{95.16±0.58}	& \small92.82±1.02	& \small91.86±1.35	& \small92.18±0.61	& \small92.69±0.7	& \small92.96±1.05\\
\cmidrule{1-1}
 {\small Autistic Spectrum}              & \small\bfseries{91.92±2.01}	& \small87.22±1.91	& \small87.15±2.56	& \small86.73±2.37	& \small87.53±1.09	& \small86.77±1.66\\
\cmidrule{1-1}
         {\small ASP-POTASSC}              & \small38.72±0.76	& \small39.99±0.92	& \small39.68±0.96	& \small36.82±0.63	& \small39.75±1.35	& \small\textbf{40.72±0.82}\\
\cmidrule{1-1}
{\small HCC survival}              & \small\bfseries{67.87±3.29}	& \small63.33±3.3	& \small64.92±2.6	& \small65.45±3.7	& \small67.57±2.76	& \small67.65±3.10\\
\cmidrule{1-1}
 {\small cervical cancer}              & \small94.5±0.5& \small	94.41±0.48	& \small94.53±0.34	& \small94.33±0.32	& \small\bfseries{94.8±0.47}	& \small94.55±0.29
\\
\cmidrule{1-1}
   {\small pbc}  & \small\textbf{64.54±1.64}	& \small64.20±1.74	& \small64.41±2.08	& \small63.21±1.42	& \small63.82±1.81	& \small64.23±1.41  \\             
\cmidrule{1-1}
   {\small horse}& \small\textbf{78.81±1.35}	& \small72.13±1.66	& \small67.86±1.51	& \small66.58±2.64	& \small70.93±1.87	& \small77.21±1.41 \\             
\cmidrule{1-1}

   {\small colic}   & \small\textbf{82.91±1.18}	& \small82.49±1.07	& \small82.61±1.33	& \small82.36±0.88	& \small81.91±1.45	& \small81.89±0.88\\                 
\cmidrule{1-1}   
   {\small AotuMLSelector}  & \small42.56±2.31	& \small48.25±2.82	& \small47.09±3.03	& \small38.67±3.61	& \small47.14±4.87	& \small\textbf{49.76±2.79}\\                    

    \hline
  \end{tabular*}
\end{table*}

\subsubsection{Results of Accuracy and $F_{1}$-score}\label{sec1}
For the real-world datasets, we use the first 5 complete datasets to simulate missing environments, and generate missing values according to MCAR and MNAR at three different proportions of 1\%, 5\%, and 10\%, respectively. In this study, all datasets undergo five-fold cross-validation. We use five baseline methods to impute the training and testing sets separately, and then use NCFS to learn feature weights on the imputed training set. Since IWMC has the functions of feature weight learning and matrix completion, we first use IWMC to learn feature weights on the training set, and then use the learned weighting vector and the loss function of the M-stage to impute the testing set. For each method's learned feature weights, we select the same number of feature items on the testing set and calculate the classification accuracy and ${F_{1}}$-score. Next, we use the remaining 9 datasets with missing values originally (Table 1) for experiments, which have different numbers of features, instances, and missing rates, representing real missing data problems.

In this work, we focus on whether imputation methods can help feature selection methods identify important features. We conduct experiments based on the classification results after feature selection because the purpose of imputing datasets is to enable downstream tasks to proceed normally, and the results of downstream tasks indirectly reflect the effectiveness of imputation. On the simulated missing environment datasets, we limit the number of selected features to 50, while on the real missing datasets, the number of selected features is 50\% of the original number of features. This allows us to see which methods can help NCFS identify the most relevant features.

From the results of the datasets with MCAR data in Tables 2 and 3, IWMC obtained the highest ${F_{1}}$-score 11 times out of 15 results for the KNN classifier, and obtained the highest classification accuracy 12 times out of 15 results for the KNN classifier. Overall, the classification accuracy and ${F_{1}}$-scores of IWMC are higher than those of the compared methods. Since the missing values are randomly generated, the variance and overall distribution of the data are almost unchanged, and the five baseline methods show similar experimental results, and mean imputation is still a strong baseline method.

Tables 4 and 5 show the results of the datasets with MNAR data. In the ${F_{1}}$-score results of Table 4, IWMC obtained the highest result 3 times less than Table 2, and in the classification accuracy results of Table 5, IWMC also obtained the highest result 3 times less than Table 3. However, in this experiment, IWMC still achieved the highest score in most of the results. We also found that under MNAR, the classification performance of most datasets is worse than that under MCAR, and the result bias is larger. This may be because when the data is MNAR, the missing values depend on the unobserved values, making the missing values harder to estimate.

On the data sets with missing values originally, As shown in Table 6, IWMC achieves the highest ${F_{1}}$-score on 7 datasets with KNN classifier, and in Table 7, IWMC achieves the highest accuracy on 6 datasets with KNN classifier. Since such datasets with missing values originally may have more outliers, SOFT imputation shows better results than other comparison methods. However, since the comparison methods did not consider the importance of features during the imputation process, their average results are worse than our method, which also confirms the effectiveness of IWMC.

\subsubsection{stability}\label{sec2}
In this work, we evaluate the stability of IWMC using boxplots and standard deviations of the accuracy on the datasets. Boxplots are a commonly used data visualization method that can show the distribution of a set of data. The box represents the interquartile range (IQR) of the data, which is the range of the middle 50\% of observations in the dataset. A longer box indicates a wider distribution of the data, while the lines extending from the box represent the maximum and minimum values of the data, including any outliers.On the other hand, the standard deviation is a statistical measure used to assess the degree of variation or dispersion of the data. A smaller standard deviation indicates that the data is closer to the mean, implying better stability.

We plotted boxplots of each algorithm's average accuracy obtained from ten runs of five-fold cross-validation on the dataset. Figures 7-8 shows the boxplots of the average accuracy of KNN classifiers on different datasets. Under MCAR, IWMC has a higher position and shorter box length compared to other methods. The results in Tables 2-3 show that in most cases, IWMC has a smaller standard deviation. In Figure 8, the results under MNAR show that IWMC does not have a significant advantage in box length, but the results in Table 4-5 show that in most cases, IWMC still has smaller bias. This could be because under MNAR, IWMC's results have a wider distribution, but the data points are not sufficiently far from the mean. Therefore, we conclude that IWMC has good imputation stability when the data is MCAR. Although its stability may decrease under MNAR, it still has a competitive advantage.

\begin{figure*}
  \centering
  \includegraphics[width=\linewidth]{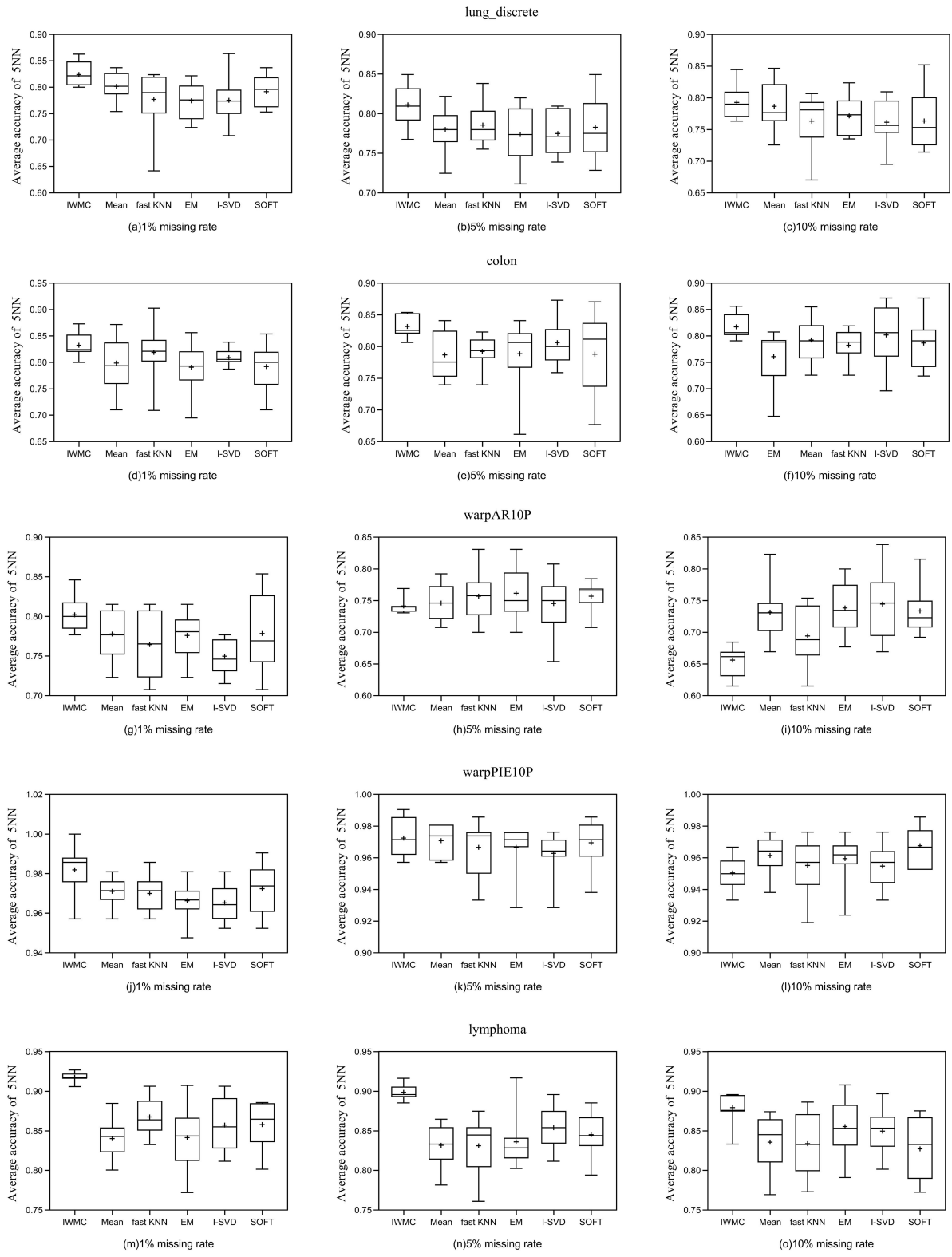}
  \caption{Boxplots of average accuracy with 5NN classifier for 5 datasets with MCAR data.}
\end{figure*}

\begin{figure*}
  \centering
  \includegraphics[width=\linewidth]{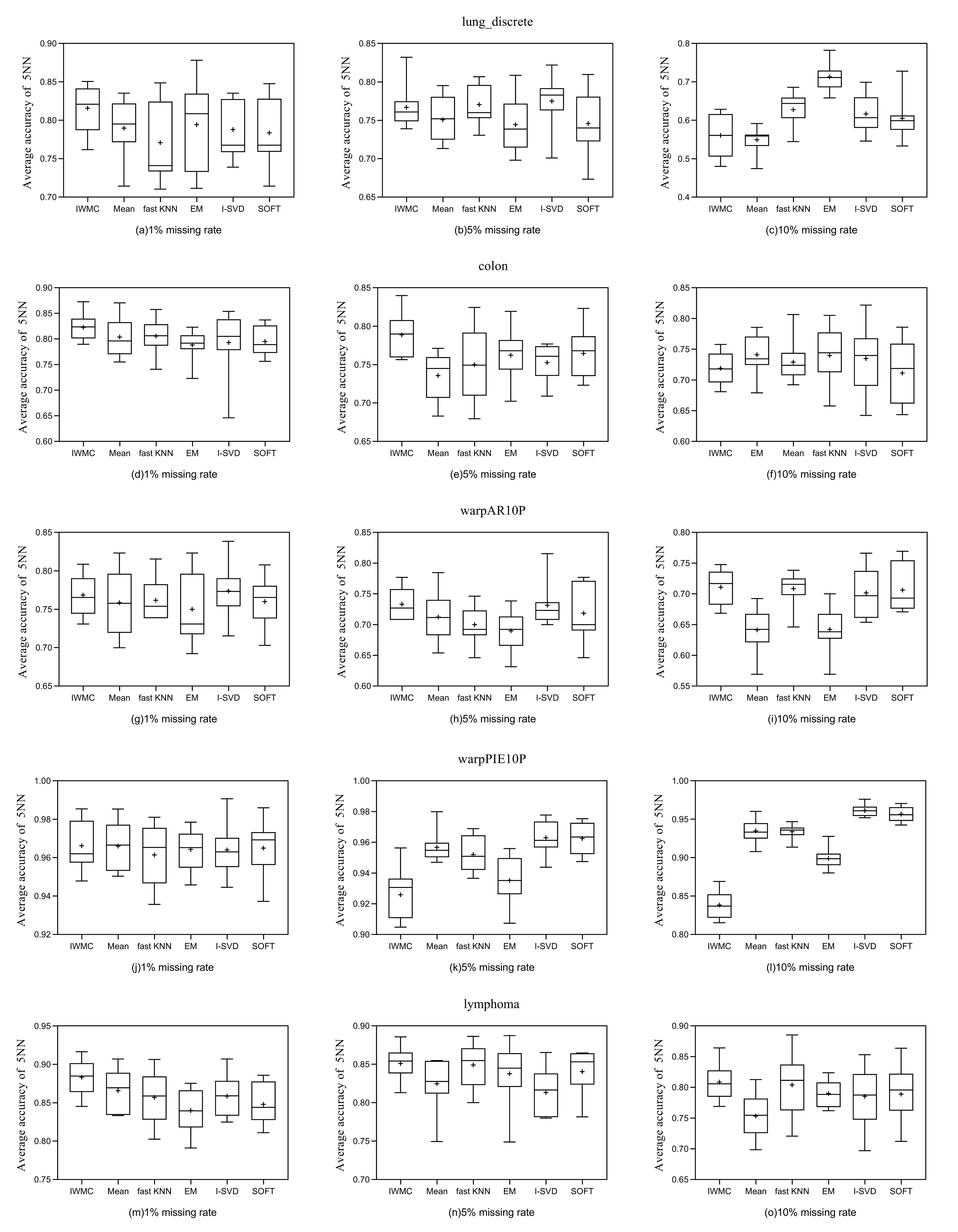}
  \caption{Boxplots of average accuracy with 5NN classifier for 5 datasets with MNAR data.}
\end{figure*}
\section{Conclusion}
Missing value imputation (MVI) is an important data preprocessing technique that effectively addresses the issue of missing values in datasets. In this paper, we propose a feature importance-based imputation method that involves two-stage iterations: the M-stage and the W-stage. The M-stage involves completing the dataset based on the current feature weighting vector {\bf{w}}, while the W-stage involves learning the weighting vector {\bf{w}} based on the completed dataset after M-stage. Numerical experiments were conducted using three types of datasets: synthetic datasets with different noise features and missing values, real datasets with different missing values, and real-world datasets originally with missing values. Experimental results comparing our proposed IWMC method to Mean imputation, EM imputation, fast KNN imputation, I-SVD, and SOFT imputation show that IWMC effectively identifies relevant features in the presence of missing values, making it an effective imputation algorithm. In the W-stage, we used the existing feature selection method NCFS to learn the feature weighting vector, but there are more advanced feature selection methods available to increase the algorithm's effectiveness, as well as feature selection methods that can handle continuous labels or unsupervised feature selection algorithms to further expand the algorithm's applicability. Finally, missing value imputation is an important research field, and there are many additional directions for further research.





\section*{Author Contributions}
Cong Guo:Writing-original draft,Conceptualization,Validation;Chun Liu:Writing review and editing;Wei Yang:Writing-review and editing,Supervision,Methodology
\section*{Declarations}
{Conflicts of Interest.}
The authors declare that they have no known competing financial interests or personal relationships that could have appeared to influence the work reported in this paper.
\section*{Availability of data and materials.}
The datasets used in this study was obtained from three publicly available repositories and they are available in the following websites:
1.http://archive.ics.uci.edu/ml/index.php
2.https://www.openml.org/ 
3.https://jundongl.github.io/scikit-feature/datasets.html


\end{document}